\documentclass[lettersize,journal]{IEEEtran}
\usepackage{amsmath,amsfonts}
\usepackage{algorithm}
\usepackage{algpseudocode}

\usepackage{array}
\usepackage{booktabs}
\usepackage{makecell}
\usepackage{multirow}
\usepackage[table]{xcolor}
\usepackage[caption=false,font=normalsize,labelfont=sf,textfont=sf]{subfig}
\usepackage{textcomp}
\usepackage{stfloats}
\usepackage{url}
\usepackage{verbatim}
\usepackage{graphicx}
\usepackage{cite}
\usepackage{subfig}

\newcommand{\rev}[1]{{#1}}
\newcommand{\revcolor}{}

\begin{document}

\title{Compact Attention: Exploiting Structured Spatio-Temporal Sparsity for Fast Video Generation}
\author{
        Qirui~Li, Guangcong~Zheng, Qi~Zhao, Jie~Li, Bin~Dong, Yiwu~Yao, Xuewei~Li, Zequn~Qin, Xi~Li$^{*}$
        
\thanks{
Qirui Li, Guangcong Zheng, Qi Zhao, Jie Li and Xi Li are with the College of Computer Science and Technology, Zhejiang University, Hangzhou 310027, China.
Xuewei~Li is with the School of Artificial Intelligence and Data Science, Shanghai Dianji University, Shanghai 201306, China.
Zequn Qin is with the School of Software Technology, Zhejiang University, Ningbo 315100, China.
Bin Dong and Yiwu Yao are with Huawei Technologies Co., Ltd., Hangzhou 310053, China.
E-mail: \{qirui.l, guangcongzheng, qizhao, li.jie, qinzequn, xilizju\}@zju.edu.cn, xueweili@sdju.edu.cn, dongbin37@huawei.com, yiwuyao@pku.edu.cn.
Qirui Li and Guangcong Zheng contributed equally to this work.
}%
\thanks{(Corresponding author: Xi~Li)}
\thanks{This work has been accepted for publication in \textit{IEEE Transactions on Multimedia}.}
\thanks{\copyright~2026 IEEE. Personal use of this material is permitted. Permission from IEEE must be obtained for all other uses, in any current or future media, including reprinting/republishing this material for advertising or promotional purposes, creating new collective works, for resale or redistribution to servers or lists, or reuse of any copyrighted component of this work in other works.}

}

\markboth{IEEE TRANSACTIONS ON MULTIMEDIA}%
{Li \MakeLowercase{\textit{et al.}}: Compact Attention: Exploiting Structured Spatio-Temporal Sparsity for Fast Video Generation}


\maketitle

\begin{abstract}

The quadratic computational complexity of self-attention mechanisms imposes a critical constraint on transformer-based video generation, particularly for synthesizing ultra-long and high-definition sequences. While most sparse attention techniques employ fixed or heuristic patterns, they often struggle to accommodate the heterogeneous and layer-specific redundancies inherent in \rev{video diffusion transformers (VDiTs)}. \rev{We find that different attention heads consistently concentrate most of their attention mass in different coarse query--key regions, a behavior we call functional specialization. Although individual attention values vary with the generated content, these regions remain similar across prompts and random seeds.} Leveraging these insights, we introduce \textbf{Compact Attention}, a hardware-aware sparse attention framework for efficient long-video synthesis. \rev{We quantify this stability and use it to construct searchable, reusable, head-specific masks.} Based on the observation, we propose a \textbf{deformable dual-window} sparse pattern mechanism that flexibly approximates local and axial attention regions within one tile-aligned mask family. To avoid manually choosing a mask for each head, we develop a \textbf{recall-driven offline auto-search} algorithm that identifies high-quality sparse masks while leveraging temporal stability for mask reuse. Extensive evaluations on Wan2.1 (14B) and Hunyuan-Video demonstrate that Compact Attention achieves \rev{2.25$\times$/3.03$\times$ attention speedups over FlashAttention-2 and 1.20$\times$/1.74$\times$ over FlashAttention-3, respectively}, while maintaining visual quality and temporal consistency comparable to full-attention baselines.
\end{abstract}


\section{Introduction}

\begin{figure*}[t!]
    \centering
    \includegraphics[width=1.0\linewidth]{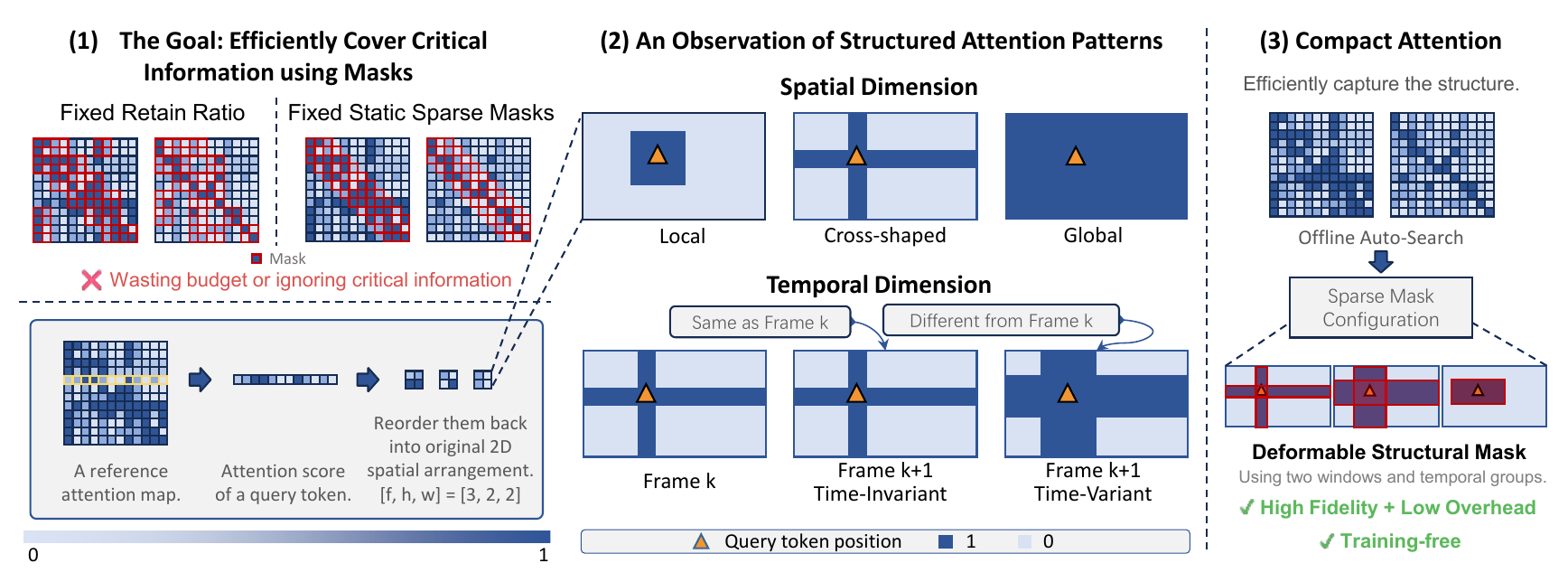}
    \caption{\rev{The motivation and overview of Compact Attention. (1) Fixed masks may waste computation or miss important attention regions; blue intensity denotes reference attention mass, while red overlays and outlines denote retained query--key regions. (2) Different heads exhibit local, axial, or dense coarse concentration regions. (3) Compact Attention uses two deformable windows and temporal-distance groups to represent the observed spatial and temporal structures.}}
    \label{fig:overview}
\end{figure*}

The rapid evolution of \rev{VDiTs} has enabled the synthesis of high-quality, photorealistic videos. Beyond incorporating complex inputs ~\cite{liu2023sounding}, ~\cite{koksal2023controllable}, ~\cite{wang2026Customvideo}, and establishing rigorous evaluation metrics ~\cite{hu2024benchmark}, generating ultra-long, high-definition sequences remains a formidable challenge due to the quadratic computational complexity of self-attention. In state-of-the-art architectures like Hunyuan-Video \cite{kong2024hunyuanvideo}, generating a 129-frame 720p video involves processing over 100K tokens, where attention computation alone consumes \rev{more than 70\%} of the total inference time. This bottleneck necessitates acceleration strategies that reduce computational overhead without sacrificing generative fidelity. \rev{Fig.~\ref{fig:overview} illustrates this motivation and the overall design of Compact Attention.}

Recent efforts to mitigate this complexity primarily focus on sparse attention~\cite{zhang2025fast}, \cite{xia2025training}, \cite{xi2025sparse}, \cite{ding2025efficient}, \cite{zhang2025spargeattn}. While methods like sliding windows~\cite{zhang2025fast} or fixed periodic patterns~\cite{xi2025sparse} offer basic acceleration, they impose rigid constraints that fail to capture heterogeneous, layer-specific redundancies~\cite{xi2025sparse}, \cite{zhang2025fast}. Furthermore, input-adaptive methods~\cite{jiang2024minference}, \cite{zhang2025spargeattn} introduce prohibitive online prediction overhead. Beyond training-free techniques, research has explored training-based schemes~\cite{xia2025training} and joint optimization like combining sparsity with quantization~\cite{liu2025fpsattention}, \cite{zhao2025paroattention} or distillation~\cite{feng2025text}, \cite{ding2025efficient}. However, many existing sparse patterns—especially fine-grained or unstructured ones—fail to align with the memory access coalescing required by modern GPUs, breaking compatibility with optimized hardware kernels. This necessitates a sparse paradigm that is inherently tile-aligned and static during inference.
\IEEEpubidadjcol

Our analysis of pretrained VDiTs, including Hunyuan-Video~\cite{kong2024hunyuanvideo} and Wan2.1~\cite{wan2025wan}, reveals \rev{that different heads consistently concentrate most of their attention mass in different coarse query--key regions, a behavior we call functional specialization}. \rev{For each primary model, we analyze 4,730 generations from 946 prompts and five random seeds per prompt.} Despite input-dependent attention values, the regions retained for the same head remain similar across these runs, which allows its binary mask to be determined offline and reused at inference.

Leveraging these insights, we introduce \textbf{Compact Attention}, a hardware-aware sparse attention framework for efficient long-video synthesis. Our key contributions are summarized as follows:

\begin{itemize}
    \item We \rev{show that binary masks for the same attention head retain similar query--key regions across inputs} in VDiTs. We provide rigorous statistical evidence \rev{using 4,730 generations for each of two primary models and cross-prompt measurements on two additional pretrained models}, enabling offline optimization.
    
    \item We propose a deformable dual-window mechanism that represents both axial corridors and local neighborhoods within one tile-aligned mask family. Coupled with a recall-driven offline search and step-wise mask reuse, it selects a mask for each head \rev{without input-dependent online search or routing}.
    
    \item Extensive evaluations on Wan2.1 (14B)~\cite{wan2025wan} and Hunyuan-Video~\cite{kong2024hunyuanvideo} demonstrate that Compact Attention achieves \rev{2.25$\times$/3.03$\times$ attention speedups over FlashAttention-2 and 1.20$\times$/1.74$\times$ over FlashAttention-3} on a single H800 GPU, while maintaining visual fidelity comparable to full-attention baselines.
\end{itemize}

\begin{figure*}[!t]
    \centering
    \includegraphics[width=1\linewidth]{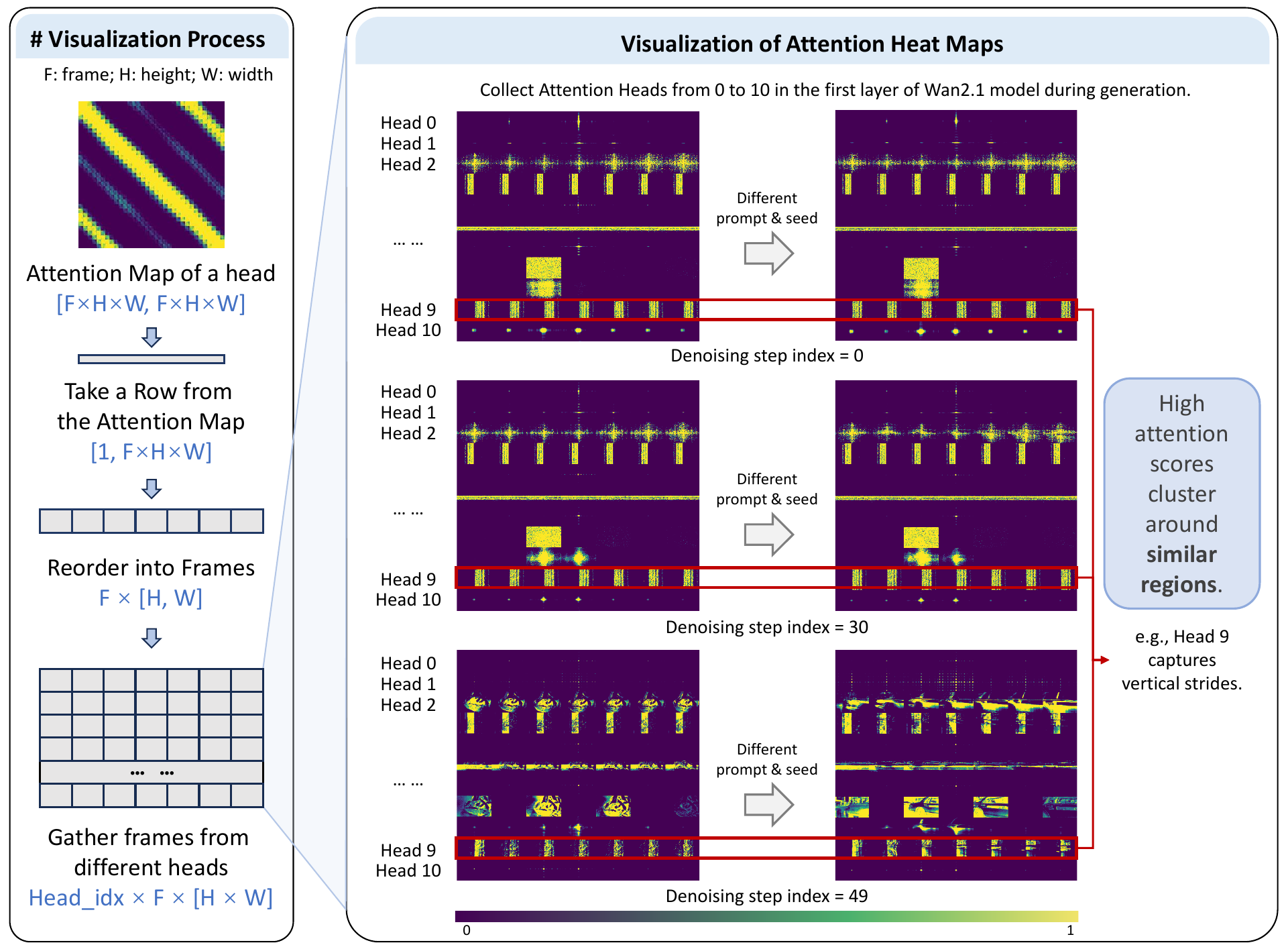}
    \caption{\rev{Coarse concentration regions of attention heads in VDiTs.} We collect attention scores from Wan2.1-14B across multiple generation runs. Query-centric attention maps, with the query token at the spatial center of the middle frame, are arranged in a film-strip layout. The two examples show a relatively static chameleon and a fast-moving car. Each column presents 11 attention heads over seven frames. Across prompts and random seeds, \rev{the same head tends to concentrate attention in similar coarse regions, while the values within those regions remain input dependent}.}
    \label{fig:visualAttn}
\end{figure*}

\begin{figure*}[t!]
    \centering
    \includegraphics[width=1\linewidth]
    {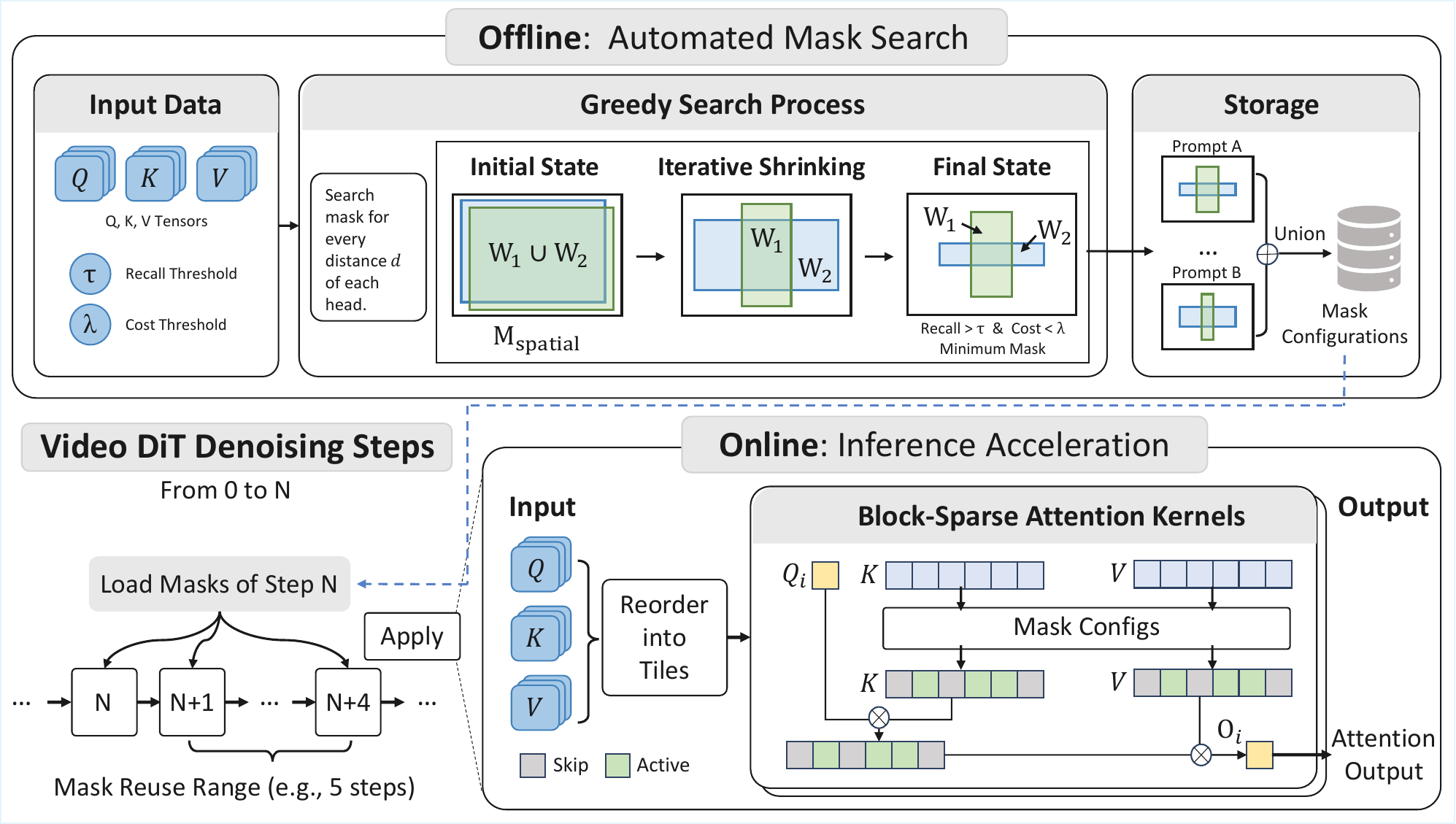}
    \caption{Overall pipeline of Compact Attention. (Top) Offline automated search iteratively shrinks dual windows under the recall ($\tau$) and cost ($\lambda$) constraints and stores the resulting sparse mask configurations. (Bottom) Online inference loads the precomputed masks into block-sparse attention kernels and reuses each configuration over a fixed interval of adjacent denoising steps.}
    \label{fig:enter-label}
    \vspace{-1.5em}
\end{figure*}

\section{Related Works}

We review work on diffusion acceleration, sparse attention, and attention-head specialization.

\subsection{Acceleration of Diffusion Models}
Diffusion acceleration often reduces denoising steps through training-free ODE/SDE solvers~\cite{song2020denoising,lu2022dpm,liu2023oms,bao2022analytic,liu2022pseudo,zhang2022fast} or training-based distillation and consistency models~\cite{salimans2022progressive,meng2023distillation,yin2024improved,li2024t2v,kim2023consistency,song2023consistency,heek2024multistep,xie2024mlcm}. A second line caches latent features or layer outputs across denoising steps~\cite{ma2024deepcache,ma2024learning,agarwal2024approximate,lv2024fastercache,kahatapitiya2024adaptive,wimbauer2024cache}. These strategies reduce function evaluations or repeated computation but do not directly reduce query--key interactions within each attention call. \rev{Attention quantization~\cite{zhao2024vidit,li2024svdqunat,zhang2024sageattention,zhang2024sageattention2} instead changes arithmetic and memory precision; Compact Attention reduces the evaluated interactions.}

\subsection{Sparse Attention and Hardware Efficiency}
Sparse attention limits token interactions. Dynamic methods identify important tokens or structures online~\cite{zhang2023h2o,StreamingLLM,xiao2024duoattention,jiang2024minference,ribar2023sparq,lai2025flexprefill,gao2024seerattention}; in video generation~\cite{zhang2025spargeattn,shen2025draftattention,yang2025sparse2}, this selection adds inference overhead. Static methods avoid online selection but commonly impose one sliding-window or grid prior across heterogeneous heads~\cite{zhang2025fast,xi2025sparse,li2025radia}. \rev{Other designs include fine-grained FG-Attn~\cite{durvasula2025fgattn}, low-resolution guidance in DraftAttention~\cite{shen2025draftattention}, trainable VSA~\cite{zhang2025vsa}, structured VMonarch~\cite{liang2026vmonarch}, and the Attention Tile used with distillation in Efficient-vDiT~\cite{ding2025efficient}. They differ in layout selection and whether model adaptation is required.}

Compact Attention bridges this gap through a deformable, tile-based dual-window design that requires \rev{no input-dependent online mask search or routing}. Its tile-aligned layout supports regular memory access in modern GPU kernels such as FlashAttention-3~\cite{shah2024flashattention}.

\subsection{Attention Head Specialization}
The viability of our offline search and static mask strategy is rooted in the functional specialization of attention heads. Here, functional specialization means that different heads consistently require different retained query--key regions. In NLP, specific heads develop roles such as induction or positional tracking~\cite{xiao2024duoattention,zhang2023h2o,fu2024moa}; in 2D Vision Transformers, attention can exhibit local or axial structures. \rev{CCNet~\cite{huang2019ccnet} imposes criss-cross connectivity by design, allowing each position to attend along its row and column, with recurrent application aggregating full-image context. By contrast, we examine recurring coarse concentration regions that emerge within dense all-to-all attention in pretrained VDiTs.}

In VDiTs, \rev{the same head often concentrates most attention mass in similar coarse query--key regions across prompts and random seeds, even as the values within those regions change}. Compact Attention uses this separation between stable retained regions and input-dependent weights to determine its masks offline.

\section{Empirical Analysis of Structural Biases in Video Diffusion Transformers}
\label{sec:analysis}

This section examines which coarse query--key regions contain most of the attention mass in pretrained VDiTs. We use \rev{functional specialization to mean that different heads consistently require different retained regions} and evaluate whether these regions remain similar across inputs and denoising steps.

\subsection{Preliminaries and Problem Formulation}
Standard self-attention processes an input sequence of video tokens $X \in \mathbb{R}^{L \times d}$, where $L$ denotes the total number of spatio-temporal tokens. The attention score matrix $A$ is computed as
\begin{equation}
    A = \mathrm{Softmax}\left(\frac{QK^\top}{\sqrt{d_k}}\right),
\end{equation}
where $Q, K \in \mathbb{R}^{L \times d_k}$ are the query and key projections. For high-definition video generation such as Hunyuan-Video, $L$ can exceed $10^5$, yielding $\mathcal{O}(L^2)$ complexity that accounts for \rev{more than 70\%} of total inference time. To improve hardware efficiency, we adopt a tile-based perspective by reorganizing tokens into spatially contiguous tiles before applying sparse masks. \rev{At tile granularity, a sparse attention mask is a binary compute layout over query--key tile pairs: active tiles execute the corresponding $QK^\top$ and $PV$ blocks, whereas masked tiles are skipped. We use \emph{attention support} only to denote this set of active query--key tiles; it specifies retained locations, not their attention values. At the same active-tile budget, a more accurate mask retains more post-softmax attention mass.}

\subsection{\rev{Cross-Input Similarity of Head-Specific Masks}}

\begin{figure*}[t!]
    \centering
    \includegraphics[width=1\linewidth]{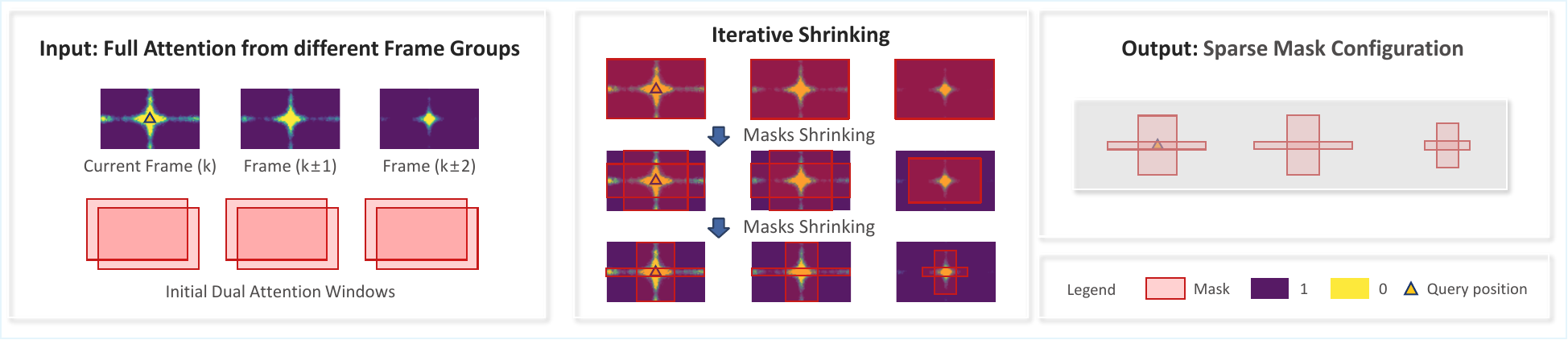}
    \caption{Detailed visualization of the offline mask auto-search process. Starting from full attention coverage across different temporal \rev{distance groups}, the dual windows are iteratively shrunk \rev{while satisfying the recall and shrink-penalty constraints}. The resulting binary mask retains the coarse query--key regions that contain most of each head's attention mass.}
    \label{fig:search}
    \vspace{-1.5em}
\end{figure*}

Although standard self-attention permits all-to-all connectivity, query-specific attention energy in the evaluated VDiTs is often concentrated within coarse geometric regions (Fig.~\ref{fig:visualAttn}). These regions include horizontal or vertical corridors and a local neighborhood around the query. Fig.~\ref{fig:search} visualizes the iterative dual-window shrinking process used to obtain the binary masks. We evaluate whether the binary masks retain similar regions across inputs, while the attention values within them continue to encode input-specific content.

To characterize cross-input stability, we compare the binary masks produced by the offline search for each attention head. \rev{For each primary model, we analyze 4,730 generations from 946 prompts and five random seeds per prompt. For every dense post-softmax attention map collected during generation, the search starts from full coverage and contracts the two windows at tile-aligned boundaries. The final window union directly defines the binary mask: retained query--key tiles are assigned 1 and skipped tiles are assigned 0. The recall threshold $\tau$ controls the minimum retained attention mass, while $\lambda$ limits the accepted recall loss per unit of computation removed.} We measure binary-mask similarity using Intersection-over-Union (IoU):
\begin{equation}
\mathrm{Sim}(M_A, M_B) = \frac{\|M_A \odot M_B\|_1}{\|M_A + M_B - M_A \odot M_B\|_1},
\end{equation}
where $M_A, M_B \in \{0,1\}^{N_q \times N_k}$ are tile-level binary masks and $\odot$ denotes element-wise multiplication. \rev{Cross-prompt IoU averages all prompt pairs at matched layers, attention heads, and temporal groups. Mean cross-prompt (cross-seed) IoU is 0.8124 (0.8274) on Wan2.1 and 0.8425 (0.8523) on Hunyuan-Video. The same protocol yields cross-prompt mean IoU values of 0.80 on Wan2.2 (Wan family~\cite{wan2025wan}) and 0.78 on CogVideoX~\cite{yang2024cogvideox}. With $\lambda$ fixed at 0.011 and 0.004 for the two primary models, the relative trend remains consistent as $\tau$ varies from 0.80 to 0.95. The recurring axial and local regions are also consistent with the 3D RoPE mechanism described for video tokens in CogVideoX~\cite{yang2024cogvideox}: its rotations make attention logits depend explicitly on relative spatio-temporal displacement, while different learned query/key projections allow individual heads to develop distinct positional responses.} Together, these results motivate determining the retained regions offline while computing input-specific attention values at inference time.

We also examine stability across denoising steps. \rev{Mask similarity between neighboring steps remains above 0.9 in our measurements, and the dominant coarse region of a head rarely switches between an axial corridor and a local neighborhood.} This persistence justifies reusing calibrated masks over short denoising-step intervals.

\section{Compact Attention}

The observations in Section~\ref{sec:analysis} show that structured sparse masks can retain the coarse query--key regions containing most attention mass in VDiTs. Because \rev{the retained regions remain similar across the evaluated calibration inputs and short denoising-step intervals}, we determine the masks offline instead of selecting them during inference. As illustrated in Fig.~\ref{fig:enter-label}, Compact Attention combines a deformable dual-window parameterization with an offline search that selects a high-recall mask \rev{for each head and temporal-distance group}.

\subsection{Deformable Dual-Window Mask Parameterization}

To make sparse attention both expressive and hardware-friendly, we define the sparse mask at the tile level. Following prior block-sparse designs~\cite{zhang2025fast}, we group tokens into contiguous spatio-temporal tiles so that the dense attention matrix is partitioned into tile-aligned blocks, where each tile contains $b = b_t \times b_h \times b_w$ tokens, which better matches the local and axial attention structures observed in practice. Sparsity is then applied by selecting which query--key tile pairs are computed or skipped, which aligns with block-sparse operations in modern GPU kernels.

To model diverse spatial patterns without inefficient manual pre-classification, we define the search space for each spatial mask as the union of two independent rectangular windows. Formally, for a query-centered spatial coordinate system, let $W_1$ and $W_2$ be two rectangular bounding boxes in the 2D key-token coordinate space. The active key region $M_{\text{spatial}}$ is defined as:
\begin{equation}
    M_{\text{spatial}} = W_1 \cup W_2.
\end{equation}
This simple yet powerful design empowers the offline search algorithm to flexibly form various shapes. It can create a standard local pattern by converging $W_1$ and $W_2$ into a single overlapping region, or naturally form a cross-shaped pattern by optimizing $W_1$ to be tall-and-thin (vertical axial bias) and $W_2$ to be short-and-wide (horizontal axial bias).

To model the observed temporal dynamics, we partition the key tiles into distance-based groups relative to the query tile. \rev{Let $G$ denote the number of temporal tile positions after VAE encoding, temporal patchification, and tiling. For a query tile at $g_q$ and a key tile at $g_k$, we define $d=|g_q-g_k|$ and assign an independent dual-window mask $M^{(d)}$ to each $d\in\{0,\ldots,G-1\}$. Thus, there are $G$ possible absolute-distance groups, and positive and negative offsets of the same magnitude share a group.} This design goes beyond a simple temporal locality assumption, which presumes that attention energy decays monotonically with distance. \rev{Long-range groups are searched independently and are not truncated a priori.}

In summary, this parameterization provides a unified and hardware-friendly search space that can express both spatial heterogeneity and temporal variation without resorting to rigid heuristic mask templates.

\begin{algorithm}
\caption{Cost-Effective Greedy Window Search}
\label{algorithm:cegws}
\small 
\begin{algorithmic}[1]
\State \textbf{Input:} Tensors $Q, K, V$, recall threshold $\tau$, shrink-penalty threshold $\lambda$, shrink set $S$.
\State \textbf{Output:} Optimized sparse mask configuration $M_{\text{best}}$.
\State $M_{\text{best}} \gets M_{\text{initial}}$; \quad $(R_{\text{last}}, C_{\text{last}}) \gets \text{Evaluate}(M_{\text{best}})$
\While{$R_{\text{last}} \ge \tau$} \Comment{Continue shrinking while recall is above threshold}
    \State $\text{Moves} \gets \emptyset$ 
    \For{each valid shrink $e \in S$}
        \State $M_{\text{cand}} \gets \text{ApplyShrinking}(M_{\text{best}}, e)$
        \State $(R_{\text{cand}}, C_{\text{cand}}) \gets \text{Evaluate}(M_{\text{cand}})$ \Comment{Compute Recall and Cost based on Eq. (4) and Eq. (5)}
        \State $\Delta R \gets R_{\text{last}} - R_{\text{cand}}$; \quad \State $\Delta C \gets C_{\text{last}} - C_{\text{cand}}$
        \If{$\Delta C > 0$} \Comment{Avoid degenerate states dynamically}
            \State $E \gets \Delta R / \Delta C$
            \State $\text{Moves} \gets \text{Moves} \cup \{(E, M_{\text{cand}}, R_{\text{cand}}, C_{\text{cand}})\}$ 
        \EndIf
    \EndFor
    \If{$\text{Moves} = \emptyset$} \textbf{break} \EndIf
    
    \State $(E_{\text{min}}, M_{\text{next}}, R_{\text{next}}, C_{\text{next}}) \gets \text{argmin}_E(\text{Moves})$ \Comment{Lowest penalty}
    \If{$E_{\text{min}} > \lambda$ \textbf{or} $R_{\text{next}} < \tau$} \textbf{break} \EndIf \Comment{Stop if thresholds exceeded}
    
    \State $M_{\text{best}} \gets M_{\text{next}}$; \quad $(R_{\text{last}}, C_{\text{last}}) \gets (R_{\text{next}}, C_{\text{next}})$
\EndWhile
\State \textbf{return} $M_{\text{best}}$
\end{algorithmic}
\end{algorithm}

\begin{table*}[!t!]
    \centering
    \renewcommand{\arraystretch}{1} 
    \setlength{\tabcolsep}{1.5mm} 
    \vspace{-2ex}
    \scriptsize 
    
    \caption{Quantitative Comparison of Sparse Attention Methods in Wan2.1 and Hunyuan on Prompts from the Open-Sora Benchmark and VBench: Visual Consistency, Aesthetic Quality, Text-Video Alignment, and Temporal Semantic Consistency.}
    \label{tab:t2v_results_vbench}
    \resizebox{\textwidth}{!}{
    \begin{tabular}{llcccccccc}
    \toprule
    Model (seq\_len) & Method & Sparsity  & \makecell{Subject\\Consistency} & \makecell{Background\\Consistency} & \makecell{Aesthetic\\Quality} & CLIPSIM & CLIP-T \\
    \midrule
    \multirow{7}{*}{\shortstack{Wan2.1 ~\cite{wan2025wan}\\(80K)}}
    & FlashAttention-2 ~\cite{dao2022flashattention} & 0.00\% & 0.9507 & 0.9623 & 0.6246 & 0.1933 & 0.9989\\
    & FlashAttention-3 ~\cite{shah2024flashattention} & 0.00\% & 0.9505 & 0.9597 & 0.6286 & 0.1943 & 0.9986\\
    \cmidrule{2-8} 
    & Sparse VideoGen1 ~\cite{xi2025sparse} & 25.02\% & {0.9487} & 0.9564 & 0.6112 & 0.1926 & \underline{0.9985}\\
    & Sparse VideoGen2 ~\cite{yang2025sparse2} & 28.10\% & 0.9436 & \underline{0.9592} & 0.6214 & \textbf{0.1942} & \textbf{0.9986}\\
    & SpargeAttn ~\cite{zhang2025spargeattn} & 32.27\% & 0.9286 & 0.9434 & 0.5898 & 0.1910 & 0.9982\\
    & {\cellcolor[rgb]{0.925,0.957,1}}\textbf{Compact Attention (Ours)} & {\cellcolor[rgb]{0.925,0.957,1}}33.99\% & {\cellcolor[rgb]{0.925,0.957,1}}\underline{0.9491} & {\cellcolor[rgb]{0.925,0.957,1}}0.9586 & {\cellcolor[rgb]{0.925,0.957,1}}\textbf{0.6219} & {\cellcolor[rgb]{0.925,0.957,1}}0.1921
    & {\cellcolor[rgb]{0.925,0.957,1}}\underline{0.9985}\\ 
    
    & {\cellcolor[rgb]{0.925,0.957,1}}\textbf{Compact Attention (Ours)} & {\cellcolor[rgb]{0.925,0.957,1}}24.66\% & {\cellcolor[rgb]{0.925,0.957,1}}\textbf{0.9495} & {\cellcolor[rgb]{0.925,0.957,1}}\textbf{0.9601} & {\cellcolor[rgb]{0.925,0.957,1}}\underline{0.6218} & {\cellcolor[rgb]{0.925,0.957,1}}\underline{0.1932} 
    & {\cellcolor[rgb]{0.925,0.957,1}}\textbf{0.9986}\\ 
    
    \midrule
    \multirow{7}{*}{\shortstack{Hunyuan ~\cite{kong2024hunyuanvideo} \\(127K)}}
    & FlashAttention-2 ~\cite{dao2022flashattention} & 0.00\% & 0.9739 & 0.9614 & 0.6642 & 0.1988 & 0.9995 \\
    & FlashAttention-3 ~\cite{shah2024flashattention} & 0.00\% & 0.9723 & 0.9615 & 0.6555 & 0.1990 & 0.9995\\
    \cmidrule{2-8}
    & Sparse VideoGen1 ~\cite{xi2025sparse} & 62.20\% & 0.9677 & 0.9584 & 0.6467 & \textbf{0.1987} & \textbf{0.9995}\\
    & Sparse VideoGen2 ~\cite{yang2025sparse2} & 58.88\% & \underline{0.9713} & \textbf{0.9602} & 0.6510 & 0.1984 & \textbf{0.9995}\\
    &  SpargeAttn ~\cite{zhang2025spargeattn} & 47.77\% & 0.9641 & {0.9577} & 0.6421 & \ 0.1983 & \underline{0.9994}\\
    
    & {\cellcolor[rgb]{0.925,0.957,1}}\textbf{Compact Attention (Ours)} & {\cellcolor[rgb]{0.925,0.957,1}}62.36\% & {\cellcolor[rgb]{0.925,0.957,1}}0.9712 & {\cellcolor[rgb]{0.925,0.957,1}}0.9586 & {\cellcolor[rgb]{0.925,0.957,1}}\underline{0.6527} & {\cellcolor[rgb]{0.925,0.957,1}}\underline{0.1986}
    & {\cellcolor[rgb]{0.925,0.957,1}}\textbf{0.9995}\\ 
    
    & {\cellcolor[rgb]{0.925,0.957,1}}\textbf{Compact Attention (Ours)} & {\cellcolor[rgb]{0.925,0.957,1}}52.90\% & {\cellcolor[rgb]{0.925,0.957,1}}\textbf{0.9724} & {\cellcolor[rgb]{0.925,0.957,1}}\underline{0.9601} & {\cellcolor[rgb]{0.925,0.957,1}}\textbf{0.6616} & {\cellcolor[rgb]{0.925,0.957,1}}\textbf{0.1987} 
    & {\cellcolor[rgb]{0.925,0.957,1}}\textbf{0.9995}\\ 
    
    \bottomrule
    \end{tabular}
    }
    \vspace{0.5em}
    \footnotesize
\end{table*}

\subsection{Offline Auto-Search with Step-Wise Reuse}
\label{sec:mask_search}

Given the mask parameterization above, we search for a sparse mask for each attention head and each frame-distance group offline. The goal is to preserve dominant attention energy while minimizing the number of active tiles, thereby avoiding the runtime cost of input-adaptive sparse routing.

\rev{For a given attention head, let $A$ denote the dense post-softmax attention probability matrix and let $M$ denote the binary tile mask expanded to the token-level shape of $A$.} We define the \textit{Recall} $R(M)$ as the proportion of attention energy preserved by the mask:
\begin{equation}
    \rev{R(M) = \frac{\sum_{i,j} M_{i,j} A_{i,j}}{\sum_{i,j} A_{i,j}}.}
\end{equation}
The \textit{Computational Cost} $C(M)$ is defined as the total number of active tiles within $M$, which directly determines the FLOPs and memory access overhead. To guide our greedy search, we evaluate the \textit{Cost-Effectiveness} (or shrink penalty) $E$ of removing a specific boundary region. It is defined as the recall loss per unit of computational cost saved:
\begin{equation}
    E = \frac{\Delta R}{\Delta C} = \frac{R(M_{\text{before}}) - R(M_{\text{after}})}{C(M_{\text{before}}) - C(M_{\text{after}})}.
\end{equation}
This metric precisely quantifies the "recall contribution per unit of mask area." A lower $E$ indicates that the region being removed contributes marginally to the overall attention energy relative to its size, making it an optimal candidate for sparsification.

Starting from full coverage, where both windows span the entire spatial plane, we optimize the mask through greedy boundary contraction, as summarized in Algorithm~\ref{algorithm:cegws}. At each iteration, the algorithm evaluates candidate shrinking actions along the horizontal or vertical boundaries of $W_1$ and $W_2$, computes the corresponding penalty $E$, and selects the valid action with the smallest penalty. Because the effective mask is defined by the union $W_1 \cup W_2$, some candidate actions may become degenerate. For example, when the two windows fully overlap, shrinking only one window may leave the union mask unchanged, resulting in $\Delta C = 0$. We therefore dynamically prune such invalid actions and only consider contractions that modify the effective active region.

Formally, each rectangular window $W_k$ ($k \in \{1, 2\}$) is parameterized by its tile-level height and width $(h_k, w_k)$, symmetrically centered at the query token. The operation ApplyShrinking in Algorithm 1 defines a discrete transition in the search space. A shrinking action $e \in S$ corresponds to reducing exactly one dimension of a specific window by a predefined tile step size $s$ (e.g., $s=2$ in our implementation to maintain tensor alignment). The updated window dimensions become:
\begin{equation}
    (h_k', w_k') = 
    \begin{cases} 
      (\max(0, h_k - s), w_k) & \text{if } e \text{ shrinks height,} \\
      (h_k, \max(0, w_k - s)) & \text{if } e \text{ shrinks width.}
    \end{cases}
    \label{eq:shrink}
\end{equation}

After the selected dimension is contracted, the new candidate mask $M_{\text{cand}}$ is reconstructed via the union formulation in Eq. (3). Since the effective mask is defined by $W_1 \cup W_2$, an independent shrink may leave the union unchanged ($\Delta C = 0$). ApplyShrinking therefore uses a geometric pruning rule. If $W_1=W_2$, both windows shrink together; if one window contains the other, only the outer window is eligible to shrink. For partially overlapping or disjoint windows, the horizontal and vertical contractions of both windows are evaluated independently. These rules ensure that every accepted action reduces the active mask area and prevent the search from revisiting unchanged masks.

\begin{table*}[!t!]
    \centering
    \renewcommand{\arraystretch}{1} 
    \setlength{\tabcolsep}{1.5mm} 
    \vspace{-2ex}
    \scriptsize 
    \caption{\rev{Quality and attention-efficiency comparison on a single H800 GPU. Attention latency is accumulated over the generation process, and speedups are reported relative to FA2 and FA3.}}
    \label{tab:t2v_results}
    \resizebox{\textwidth}{!}{
    \begin{tabular}{llccccccc}
    \toprule
    \multirow{2}{*}{Model (seq\_len)} & \multirow{2}{*}{Method} & \multirow{2}{*}{Sparsity} & \multicolumn{3}{c}{Quality} & \multicolumn{3}{c}{\rev{Attention Efficiency}} \\
    \cmidrule(lr){4-6} \cmidrule(l){7-9}
     & & & SSIM $\uparrow$ & PSNR $\uparrow$ & MSE $\downarrow$ & \rev{Attention Latency (s)} & \rev{vs. FA2} & \rev{vs. FA3} \\ 
    \midrule
    \multirow{7}{*}{\shortstack{Wan2.1 ~\cite{wan2025wan}\\(80K)}}
    & FlashAttention-2 ~\cite{dao2022flashattention} & 0.00\% & - & - & - & 1496.7s& 1.00x & \rev{0.53x} \\
    
    & FlashAttention-3 ~\cite{shah2024flashattention} & 0.00\% & - & - & - & 795.63s& 1.88x & \rev{1.00x} \\
    
    \cmidrule{2-9} 
    
    & Sparse VideoGen1 ~\cite{xi2025sparse}& 25.02\%& 0.7115 & 21.1444 & 0.0109 & 1107.9s & 1.35x & \rev{0.72x}\\

    & Sparse VideoGen2 ~\cite{yang2025sparse2}& 28.10\%& 0.7428 & 22.0998 & 0.0094 & 1118.8s & 1.34x & \rev{0.71x}\\

    & SpargeAttn ~\cite{zhang2025spargeattn}& 32.27\%& 0.6767 & 20.1787 & 0.0120 & 1065.8s & 1.40x & \rev{0.75x}\\
    
    & {\cellcolor[rgb]{0.925,0.957,1}}\textbf{Compact Attention (Ours)} & {\cellcolor[rgb]{0.925,0.957,1}}33.99\% & {\cellcolor[rgb]{0.925,0.957,1}}\underline{0.7650} & {\cellcolor[rgb]{0.925,0.957,1}}\underline{22.9692} & {\cellcolor[rgb]{0.925,0.957,1}}\underline{0.0090} & {\cellcolor[rgb]{0.925,0.957,1}}\textbf{663.82s} & {\cellcolor[rgb]{0.925,0.957,1}}\textbf{2.25x} & {\cellcolor[rgb]{0.925,0.957,1}}\rev{\textbf{1.20x}} \\
    
    & {\cellcolor[rgb]{0.925,0.957,1}}\textbf{Compact Attention (Ours)} & {\cellcolor[rgb]{0.925,0.957,1}}24.66\% & {\cellcolor[rgb]{0.925,0.957,1}}\textbf{0.7897} & {\cellcolor[rgb]{0.925,0.957,1}}\textbf{23.6788} & {\cellcolor[rgb]{0.925,0.957,1}}\textbf{0.0082} & {\cellcolor[rgb]{0.925,0.957,1}}\underline{758.18s} & {\cellcolor[rgb]{0.925,0.957,1}}\underline{1.97x} & {\cellcolor[rgb]{0.925,0.957,1}}\rev{\underline{1.05x}} \\
        \midrule
        
    \multirow{5}{*}{\shortstack{Wan2.1 ~\cite{wan2025wan}\\(69K)}}
    & FlashAttention-2 ~\cite{dao2022flashattention} & 0.00\% & - & - & - & 1106.0s& 1.00x & \rev{0.52x} \\
    
    & FlashAttention-3 ~\cite{shah2024flashattention} & 0.00\% & - & - & - & 580.44s& 1.91x & \rev{1.00x} \\

     \cmidrule{2-9} 
    
    & STA ~\cite{zhang2025fast}& 32.67\%& 0.7510 & 21.2516 & 0.0101 & \textbf{456.36s} & \textbf{2.42x} & \rev{\textbf{1.27x}}\\
    
    & {\cellcolor[rgb]{0.925,0.957,1}}\textbf{Compact Attention (Ours)} & {\cellcolor[rgb]{0.925,0.957,1}}33.99\% & {\cellcolor[rgb]{0.925,0.957,1}}\underline{0.7754} & {\cellcolor[rgb]{0.925,0.957,1}}\underline{22.9784} & {\cellcolor[rgb]{0.925,0.957,1}}\underline{0.0085} & {\cellcolor[rgb]{0.925,0.957,1}}\underline{481.86s} & {\cellcolor[rgb]{0.925,0.957,1}}\underline{2.30x} & {\cellcolor[rgb]{0.925,0.957,1}}\rev{\underline{1.20x}} \\
    
    & {\cellcolor[rgb]{0.925,0.957,1}}\textbf{Compact Attention (Ours)} & {\cellcolor[rgb]{0.925,0.957,1}}24.66\% & {\cellcolor[rgb]{0.925,0.957,1}}\textbf{0.7989} & {\cellcolor[rgb]{0.925,0.957,1}}\textbf{24.6743} & {\cellcolor[rgb]{0.925,0.957,1}}\textbf{0.0079} & {\cellcolor[rgb]{0.925,0.957,1}}{539.86s} & {\cellcolor[rgb]{0.925,0.957,1}}{2.05x} & {\cellcolor[rgb]{0.925,0.957,1}}\rev{1.08x} \\
        \midrule
    \multirow{7}{*}{\shortstack{Hunyuan ~\cite{kong2024hunyuanvideo} \\(127K)}}
    & FlashAttention-2 ~\cite{dao2022flashattention} & 0.00\% & - & - & - & 1655.2s& 1.00x & \rev{0.57x} \\
    
    & FlashAttention-3 ~\cite{shah2024flashattention} & 0.00\% & - & - & - & 951.32s& 1.74x & \rev{1.00x} \\

     \cmidrule{2-9} 

    & Sparse VideoGen1 ~\cite{xi2025sparse}& 62.20\%& 0.8023 & 24.8463 & 0.0023 & \underline{633.62s} & \underline{2.61x} & \rev{\underline{1.50x}}\\

    & Sparse VideoGen2 ~\cite{yang2025sparse2}& 58.88\%& \underline{0.8674} & \underline{27.5385} & \underline{0.0012} & 724.58s & {2.28x} & \rev{1.31x}\\

    & SpargeAttn ~\cite{zhang2025spargeattn}& 47.77\%& 0.7867 & 23.1763 & 0.0032 & 1148.6s & 1.44x & \rev{0.83x}\\
    
    & {\cellcolor[rgb]{0.925,0.957,1}}\textbf{Compact Attention (Ours)} & {\cellcolor[rgb]{0.925,0.957,1}}62.36\% & {\cellcolor[rgb]{0.925,0.957,1}}0.8662 & {\cellcolor[rgb]{0.925,0.957,1}}27.4810 & {\cellcolor[rgb]{0.925,0.957,1}}0.0013 & {\cellcolor[rgb]{0.925,0.957,1}}\textbf{546.50s}& {\cellcolor[rgb]{0.925,0.957,1}}\textbf{3.03x} & {\cellcolor[rgb]{0.925,0.957,1}}\rev{\textbf{1.74x}} \\
    
    & {\cellcolor[rgb]{0.925,0.957,1}}\textbf{Compact Attention (Ours)} & {\cellcolor[rgb]{0.925,0.957,1}}52.90\% & {\cellcolor[rgb]{0.925,0.957,1}}\textbf{0.9028} & {\cellcolor[rgb]{0.925,0.957,1}}\textbf{30.2648} & {\cellcolor[rgb]{0.925,0.957,1}}\textbf{0.0009} & {\cellcolor[rgb]{0.925,0.957,1}}{750.20s} & {\cellcolor[rgb]{0.925,0.957,1}}2.21x & {\cellcolor[rgb]{0.925,0.957,1}}\rev{1.27x} \\
        \midrule
    \multirow{5}{*}{\shortstack{Hunyuan ~\cite{kong2024hunyuanvideo} \\(115K)}}
    & FlashAttention-2 ~\cite{dao2022flashattention} & 0.00\% & - & - & - & 1304.0s& 1.00x & \rev{0.56x} \\
    
    & FlashAttention-3 ~\cite{shah2024flashattention} & 0.00\% & - & - & - & 723.93s& 1.80x & \rev{1.00x} \\

     \cmidrule{2-9} 

    & STA ~\cite{zhang2025fast}& 58.37\%& 0.8363 & 26.8853 & 0.0016 & \textbf{413.96s} & \textbf{3.15x} & \rev{\textbf{1.75x}}\\
    
    & {\cellcolor[rgb]{0.925,0.957,1}}\textbf{Compact Attention (Ours)} & {\cellcolor[rgb]{0.925,0.957,1}}62.36\% & {\cellcolor[rgb]{0.925,0.957,1}}\underline{0.9015} & {\cellcolor[rgb]{0.925,0.957,1}}\underline{29.9375} & {\cellcolor[rgb]{0.925,0.957,1}}\underline{0.0010} & {\cellcolor[rgb]{0.925,0.957,1}}\underline{476.45s}& {\cellcolor[rgb]{0.925,0.957,1}}\underline{2.74x} & {\cellcolor[rgb]{0.925,0.957,1}}\rev{\underline{1.52x}} \\
    
    & {\cellcolor[rgb]{0.925,0.957,1}}\textbf{Compact Attention (Ours)} & {\cellcolor[rgb]{0.925,0.957,1}}52.90\% & {\cellcolor[rgb]{0.925,0.957,1}}\textbf{0.9427} & {\cellcolor[rgb]{0.925,0.957,1}}\textbf{34.2755} & {\cellcolor[rgb]{0.925,0.957,1}}\textbf{0.0004} & {\cellcolor[rgb]{0.925,0.957,1}}{548.18s} & {\cellcolor[rgb]{0.925,0.957,1}}{2.38x} & {\cellcolor[rgb]{0.925,0.957,1}}\rev{1.32x}
    \\
    \bottomrule
    \end{tabular}
}
\end{table*}

The search uses a minimum recall threshold $\tau$ and a maximum shrink-penalty threshold $\lambda$. \rev{For each model, we search on four calibration prompts and aggregate the prompt-specific masks by union, retaining every connection selected for at least one prompt. As the calibration set grows, this union mask retains more connections: quality improves but effective sparsity decreases. Our 1--16 prompt sensitivity study therefore uses four prompts as the quality--sparsity trade-off for the main experiments. At a fixed size of four prompts, four independently sampled calibration sets yield sparsity of $34.3\pm1.6\%$ and mask IoU of $0.893\pm0.017$ on Wan2.1, and $62.4\pm2.4\%$ and $0.908\pm0.017$ on Hunyuan-Video.}

In addition, we exploit temporal stability across denoising steps during mask search. Since neighboring denoising steps exhibit highly similar sparse layouts, it is unnecessary to search a new mask at every step. Instead, we perform mask search once every $n$ denoising steps and reuse the searched configuration within the interval. This effectively defines a step-wise search schedule over the denoising trajectory. In our implementation, we update the sparse masks every five denoising steps and reuse the mask from the first step for the remaining steps in the interval. \rev{This schedule operates along denoising iterations, independently of the temporal-distance grouping within each attention map.}

\section{Experiments}
\label{sec:exp}

\subsection{Experimental Setup}
\label{sec:setup}

\begin{figure*}[!t]
    \centering
    \includegraphics[width=1\linewidth]{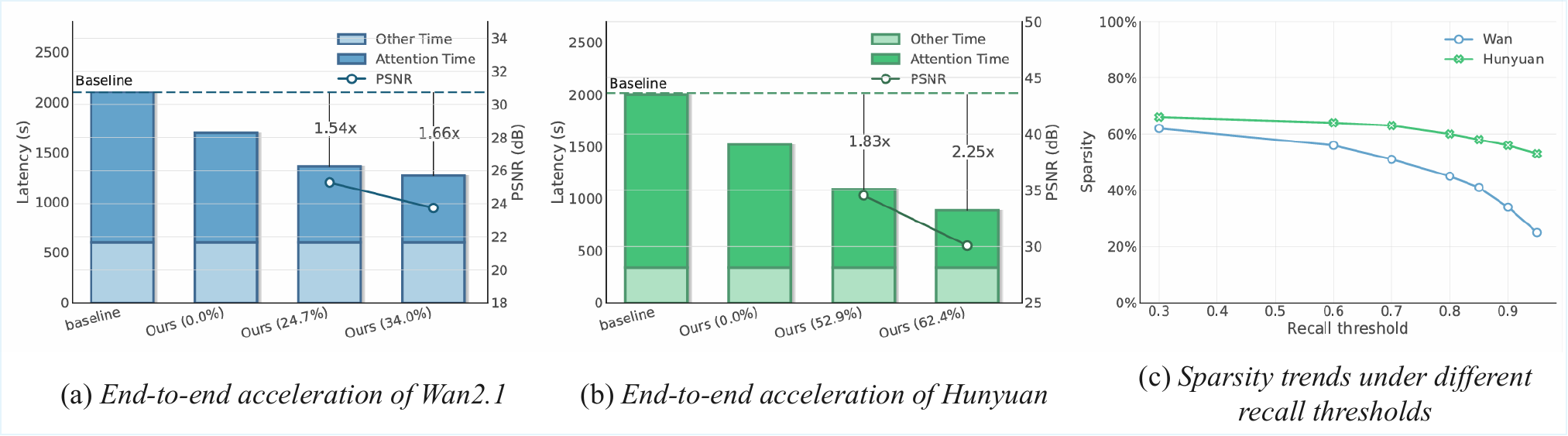}
    \caption{End-to-end performance and sparsity trends. (a-b) Latency breakdown and PSNR evaluation on Wan2.1 and Hunyuan, showing notable speedups with minimal quality degradation. Here, ``baseline'' denotes FA2, and ``ours (0.0\%)'' denotes the dense Compact kernel. (c) The impact of different recall thresholds on the achieved sparsity for auto-searched attention patterns in both models.}
    \label{fig:PSNR_distribution}
\end{figure*}

\noindent\parbox{\columnwidth}{\rev{The full-length 81/129-frame experiments use tile shape $(3,8,8)$; the 69/117-frame STA-compatible comparisons use $(6,8,8)$.}}\par
Our \rev{online inference} evaluations are conducted on two state-of-the-art video generation architectures, Wan2.1 (14B)~\cite{wan2025wan} and Hunyuan-Video~\cite{kong2024hunyuanvideo}, on a single H800 GPU. We apply Compact Attention to generate videos of 81 frames for Wan2.1 (seq\_len = 80k) and 129 frames for Hunyuan-Video (seq\_len = 127k) at a resolution of 768 $\times$ 1280. To evaluate acceleration efficacy and visual fidelity, we measure SSIM, PSNR, MSE, three VBench metrics (subject consistency, background consistency, and aesthetic quality), text-video alignment using CLIPSIM, and temporal semantic consistency using CLIPTemp (CLIP-T)~\cite{liu2024evalcrafter} on the Open-Sora benchmark. \rev{Effective sparsity is the fraction of attention computation skipped over the complete denoising run. PSNR is averaged over evaluation prompt--seed pairs, with each sparse output compared against the Full Attention output generated from the same prompt and seed. For both models, we use the union of masks searched from four calibration prompts. Here, calibration denotes this one-time offline mask search. It runs on eight H800 GPUs: Wan2.1 takes 0.73 hours (5.8 GPU-hours), with 72 GB peak memory per GPU and a 2.8 MB mask, while Hunyuan-Video takes 0.86 hours (6.9 GPU-hours), with 76 GB per GPU and a 4.2 MB mask. Calibration is excluded from per-video inference latency; Supplementary Table~S-4 reports calibration sensitivity.} Compact Attention is implemented using ThunderKittens and FlashAttention-3, with reference to the STA framework.

\textbf{Baselines:} We evaluate Compact Attention against STA~\cite{zhang2025fast}, Sparse VideoGen~\cite{xi2025sparse}, Sparse VideoGen2~\cite{yang2025sparse2}, and SpargeAttn~\cite{zhang2025spargeattn}. For a fair comparison, each baseline operates at a sparsity level similar to, or lower than, that of our method. We measure similarity relative to Full Attention and report attention speedup relative to both FlashAttention-2~\cite{dao2022flashattention} and FlashAttention-3~\cite{shah2024flashattention}. \rev{Supplementary Tables~S-1--S-3 provide additional controlled comparisons.}

\subsection{Acceleration Performance and Quality Preservation} 

\begin{figure*}[t]
    \centering
    \includegraphics[width=1\linewidth]{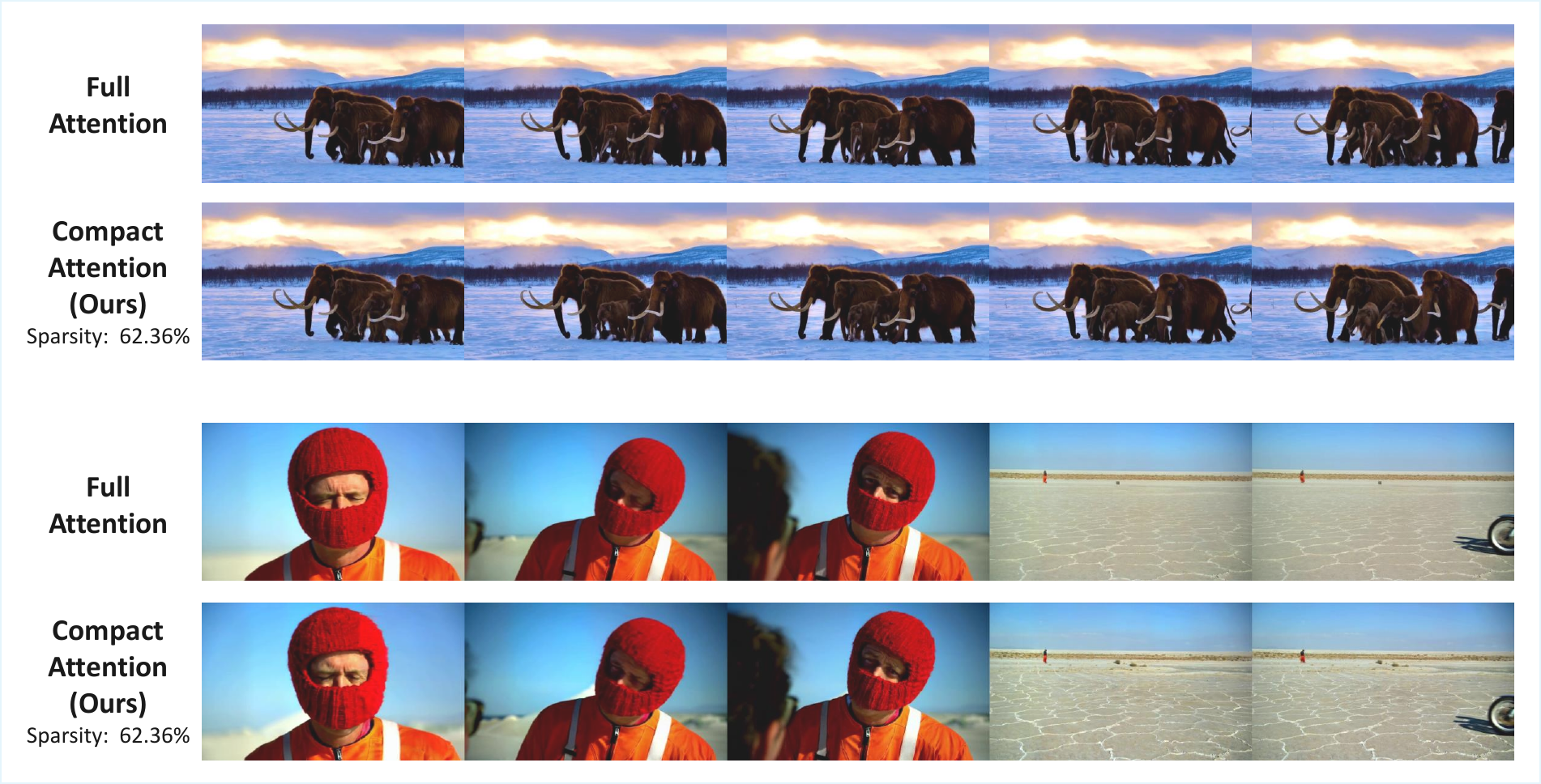}
    \caption{ Visual qualitative comparison on the Hunyuan-Video model. We compare the generated frames using Full Attention against Compact Attention at a sparsity level of 62.36\%. As shown, Compact Attention preserves the semantic details and temporal consistency (e.g., facial features and background structures) while skipping over half of the attention computation.}
    \label{fig:visual_performance}
    \vspace{-1.2em}
\end{figure*}

\paragraph{Quality Evaluation}
Table~\ref{tab:t2v_results_vbench} reports VBench quality, while Figs.~\ref{fig:PSNR_distribution} and \ref{fig:visual_performance} show the latency/PSNR and visual comparisons. \rev{Measured end-to-end FA3 speedups are 1.10$\times$/1.44$\times$ on Wan2.1/Hunyuan-Video; Supplementary Table~S-6 reports the complete wall-clock measurements.}

\paragraph{Similarity Evaluation} 
Because generating videos below the recommended resolution may degrade quality, and STA operates at constrained resolutions, we additionally compare video similarity with Full Attention using 69 frames for Wan2.1 and 117 frames for Hunyuan-Video. \rev{Both models use 4$\times$ temporal VAE reduction, temporal patch size $p_t=1$, and tile shape $(6,8,8)$ in this comparison. The resulting 18 and 30 patched latent temporal positions form three and five temporal tiles, respectively, and hence three and five possible absolute-distance groups. Supplementary Table~S-5 ablates tile granularity.} Table~\ref{tab:t2v_results} reports the quality and attention efficiency of Compact Attention and the baselines. At 62.36\% sparsity on Hunyuan-Video, Compact Attention obtains 29.9375 dB PSNR in the STA-compatible setting.

\begin{table*}[!t]
\centering
\caption{Ablation study on the sparse mask design. Components are added \textbf{cumulatively}, progressively increasing the achievable sparsity. Category-wise and overall sparsity are reported as percentages. Values in parentheses indicate the \textbf{incremental increase in percentage points (pp)} from the previous row.}
\label{tab:ablation_window_design}
\resizebox{\linewidth}{!}{%
\begin{tabular}{ccc|ccc|cc|c}
\toprule
\multicolumn{3}{c|}{\textbf{Mask Components}} & \multicolumn{3}{c|}{\textbf{Spatial-Pattern Sparsity}} & \multicolumn{2}{c|}{\textbf{Temporal-Bias Sparsity}} & \multirow{2}{*}{\textbf{Overall Sparsity}} \\
\cmidrule(r){1-3} \cmidrule(lr){4-6} \cmidrule(l){7-8}
\textbf{Cubic} & \textbf{Frame-group} & \textbf{Dual Window} & \textbf{Locality} & \textbf{Cross} & \textbf{Global} & \textbf{Time-Variant} & \textbf{Time-Invariant} & \\
\midrule
\checkmark & & & 72.6\% & 38.5\% & 7.8\% & 44.1\% & 30.6\% & 36.1\% \\
\checkmark & \checkmark & & 75.8\% & 40.6\% & 8.5\% & 47.2\% & 31.7\% & 37.0\% \textbf{(+0.9 pp)} \\
\checkmark & \checkmark & \checkmark & 76.6\% & 51.6\% & 9.9\% & 56.7\% & 38.5\% & 45.9\% \textbf{(+8.9 pp)} \\
\bottomrule
\end{tabular}%
}
\end{table*}

\subsection{Ablation Studies}
\label{sec:ablation}

\paragraph{Effectiveness of deformable mask design} 
To validate the tile-based mask design, we conduct ablations with $\tau=0.9$ and $\lambda=0.011$ on Wan2.1 (Table~\ref{tab:ablation_window_design}). Starting from a cubic window, we add frame-group-wise configurations and then dual windows. The frame-group-wise configuration increases overall sparsity by 0.9 percentage points, and the dual-window parameterization provides a further 8.9 percentage points.

\paragraph{Role of Head-Specific Masks}
\rev{Table~\ref{tab:mask_controls} compares the selected head-specific configurations with three alternatives at identical per-layer active-tile budgets: sharing one mask within a layer, permuting masks across heads, and sampling masks from the same dual-window search space. The random search also preserves the active-tile allocation across temporal-distance groups. The selected configurations improve mean PSNR by 3.85--4.66~dB on Wan2.1 and 4.62--7.03~dB on Hunyuan. This consistent gap shows that the result does not arise solely from the flexibility of the mask family; we do not attribute it to a single design factor in isolation.}
\begin{table}[!t]
\revcolor
\centering
\caption{Head-specific mask assignment with matched per-layer budgets. Mean PSNR is reported.}
\label{tab:mask_controls}
\footnotesize
\renewcommand{\arraystretch}{1.02}
\begin{tabular*}{\columnwidth}{@{\extracolsep{\fill}}llrr@{}}
\toprule
Model & Assignment & Sparsity & PSNR \\
\midrule
Wan2.1 & Correct per-head & 33.99\% & \textbf{22.97} \\
Wan2.1 & Shared within layer & 33.99\% & 18.65 \\
Wan2.1 & Random head assignment & 33.99\% & 18.31 \\
Wan2.1 & Random Mask Search & 33.99\% & 19.12 \\
\midrule
Hunyuan & Correct per-head & 62.36\% & \textbf{27.48} \\
Hunyuan & Shared within layer & 62.36\% & 22.86 \\
Hunyuan & Random head assignment & 62.36\% & 21.94 \\
Hunyuan & Random Mask Search & 62.36\% & 20.45 \\
\bottomrule
\end{tabular*}
\end{table}

\paragraph{Greedy Search Strategy}

To demonstrate the necessity of greedy search, we compare it against searching from a small, pre-selected set of 15 frequent fixed mask configurations (method used in STA). \rev{Our greedy search consistently finds configurations with both higher sparsity and higher PSNR than the pre-selected set on Wan2.1 and Hunyuan.} By tailoring the mask shape to each head's distribution, the search explores configurations that are unavailable to the fixed template set.

\paragraph{Sensitivity to Recall Threshold in Auto-Search}
Our auto-search trade-off is governed by the recall threshold ($\tau$) and cost threshold ($\lambda$). Fig.~\ref{fig:PSNR_distribution} illustrates this on both models. A higher recall threshold (e.g., 0.95) preserves more connections, leading to lower sparsity but ensuring higher fidelity. Conversely, relaxing the threshold to 0.9 allows for significantly greater acceleration.

\paragraph{Sensitivity to Sparsification in Early Denoising Steps}

Generated videos are sensitive to sparsification during early, high-noise denoising stages. Keeping Full Attention for the first rather than the final 15 steps improves PSNR by 1.02~dB; Compact Attention is therefore applied only after these initial steps. \rev{Supplementary Fig.~S-1 reports the complete schedule ablation, visual comparison, and latency.}

\section{Conclusion}
\label{sec:conclusion}

Compact Attention uses reusable, head-specific dual-window masks found by an offline search to reduce attention computation in VDiTs. \rev{This training-free design avoids input-dependent mask selection during inference. Across four pretrained models, binary masks for the same head retain similar coarse query--key regions across inputs. On Wan2.1 and Hunyuan-Video, Compact Attention achieves attention speedups of 2.25$\times$/3.03$\times$ over FA2 and 1.20$\times$/1.74$\times$ over FA3, respectively, while preserving generation quality.}

\rev{Compact Attention requires a one-time calibration search for each model checkpoint and spatio-temporal token-grid configuration. Once the checkpoint and deployment configuration are fixed, the searched patterns remain fixed and can be reused across subsequent generations, amortizing the calibration cost over repeated inference. Extending pattern transfer across checkpoints and token-grid configurations remains an important direction for future work.}

\bibliographystyle{IEEEtran}
\bibliography{IEEEabrv,main}

\end{document}